\documentclass[letterpaper, 10 pt, conference]{ieeeconf}  

\IEEEoverridecommandlockouts                              

\usepackage{amsmath,amsfonts,amssymb,amscd, fancyhdr, color, comment, graphicx, environ}
\usepackage{algorithm}
\usepackage{algpseudocode}
\usepackage{censor}
\usepackage{subcaption}
\usepackage{booktabs}
\usepackage{multirow}
\usepackage{xcolor}
\usepackage[hidelinks]{hyperref}
\title{\LARGE \bf
Embodied Active Learning under Limited Annotation and Navigation Budget for Object Detection 
}

\author{Hadrien Crassous$^{1,*}$, Mohamed Yassine Kabouri$^{2,*}$, Minahil Raza$^{3,*}$, Joni Pajarinen$^{3}$, Riad Akrour$^{1}$ %
    \thanks{$^{*}$Equal contribution.}%
    \thanks{$^{1}$ Hadrien Crassous and Riad Akrour are with Inria Scool, the Université de Lille and also with the Centre de Recherche en Informatique, Signal et Automatique de Lille (CRIStAL) (e-mail: hadrien.crassous@inria.fr and riad.akrour@inria.fr)}%
    \thanks{$^{2}$ Mohamed Yassine Kabouri is with LAAS-CNRS, Universite de Toulouse, CNRS, Toulouse, France (email: mykabouri@laas.fr)}%
    \thanks{$^{3}$ Minahil Raza and Joni Pajarinen are with the Department of Electrical Engineering and Automation at Aalto University (e-mail: minahil.raza@aalto.fi and joni.pajarinen@aalto.fi)} %
    \thanks{Correspondence to: Hadrien Crassous (hadrien.crassous@inria.fr)}
}

\begin{document}

\newpage

\maketitle
\thispagestyle{empty}
\pagestyle{empty}

\begin{abstract}
    This paper studies how to adapt a computer vision object detector to an unknown environment under both a robot navigation time and annotation budget constraint. Our approach selects informative robot trajectories and image samples to retrain the detector, explicitly targeting its failure cases. Formally, the approach is an embodied variant of batch active learning, where at each round an agent has a limited navigation budget to collect candidate samples and a limited annotation budget for the most relevant images. 
    We leverage spatial consistency to identify images with inconsistent labels, which are likely to provide the greatest improvement to the vision model.
    We evaluate the approach using different active learning objectives on large scenes from the AI2-THOR simulator and on a real-world setup using a Boston Dynamics Spot robot with the real-time object detector YOLOv5. 
    Through comparison against several baselines, our experimental results show that spatial inconsistency helps guide the agent and select relevant images without external supervision, achieving the highest detection accuracy at the end of the adaptation process under the same budget. The open-source project can be found at: \textcolor{blue}{\url{https://mkabouri.github.io/embodied-active-learning-od/} }
\end{abstract}
\section{Introduction}\label{sec:introduction}
Robust object detection (OD) \cite{zhao_object_2019} is fundamental for mobile robots performing downstream tasks such as Object Search \cite{zheng_system_2023, zheng_generalized_2023}, Embodied Question Answering \cite{gordon_iqa_2018, Majumdar_2024_CVPR} and Instruction Following \cite{pmlr-v205-li23a, wu_embodied_2025}.

Embodied AIs generally rely on real-time detectors \cite{munoz_embedded_2025, hong_education_2024, zheng_towards_2022, redmon_you_2016, jocher_ultralyticsyolov5_2022, zhao_detrs_2024}, typically pretrained on standard datasets like COCO \cite{lin_microsoft_2015} or VOC2017  \cite{everingham_pascal_2010} and fine-tuned on a fixed set of labeled examples representing the task domain. 
Finetuning object detectors is an expensive process due to human annotation availability. Recent vision foundation models trained on internet-scale data \cite{zhou_detecting_2022, liu_grounding_2025, zhai_scaling_2022} are computationally expensive and are rarely deployed on robot hardware for real-time use. While these models can serve as powerful oracles for generating large-scale pseudo-labels, they are not perfect—especially on small, rare, or domain-specific objects that are less common in their training data. In our physical robot experiments, we leverage foundation models to bootstrap the labeling process by automatically annotating data, and then we selectively correct their errors through human intervention.

Although many models score well on benchmarks, real-world object detection remains difficult. Object detectors often generalize poorly and suffer from domain shift \cite{khodabandeh_robust_2019, Chen_2018_CVPR}. Errors in object detection is a prevalent failure mode in instruction-following and object search tasks \cite{wu_embodied_2025, zheng_towards_2022, zheng_generalized_2023}. Models also struggle with underrepresented classes and environmental variations such as lighting changes, texture differences, and novel spatial layouts.
Developing systems that autonomously adapt through interactions when deployed in new environments is therefore an important research problem \cite{maiettini_interactive_2017, chaplot_semantic_2020,  pmlr-v270-mirchandani25a, shi_embodied_2025}. We focus on bridging the gap between non-interactive active learning and embodied active learning, where robots actively explore and collect informative training samples.

\begin{figure}
    \centering
    \includegraphics[width=0.5\textwidth]{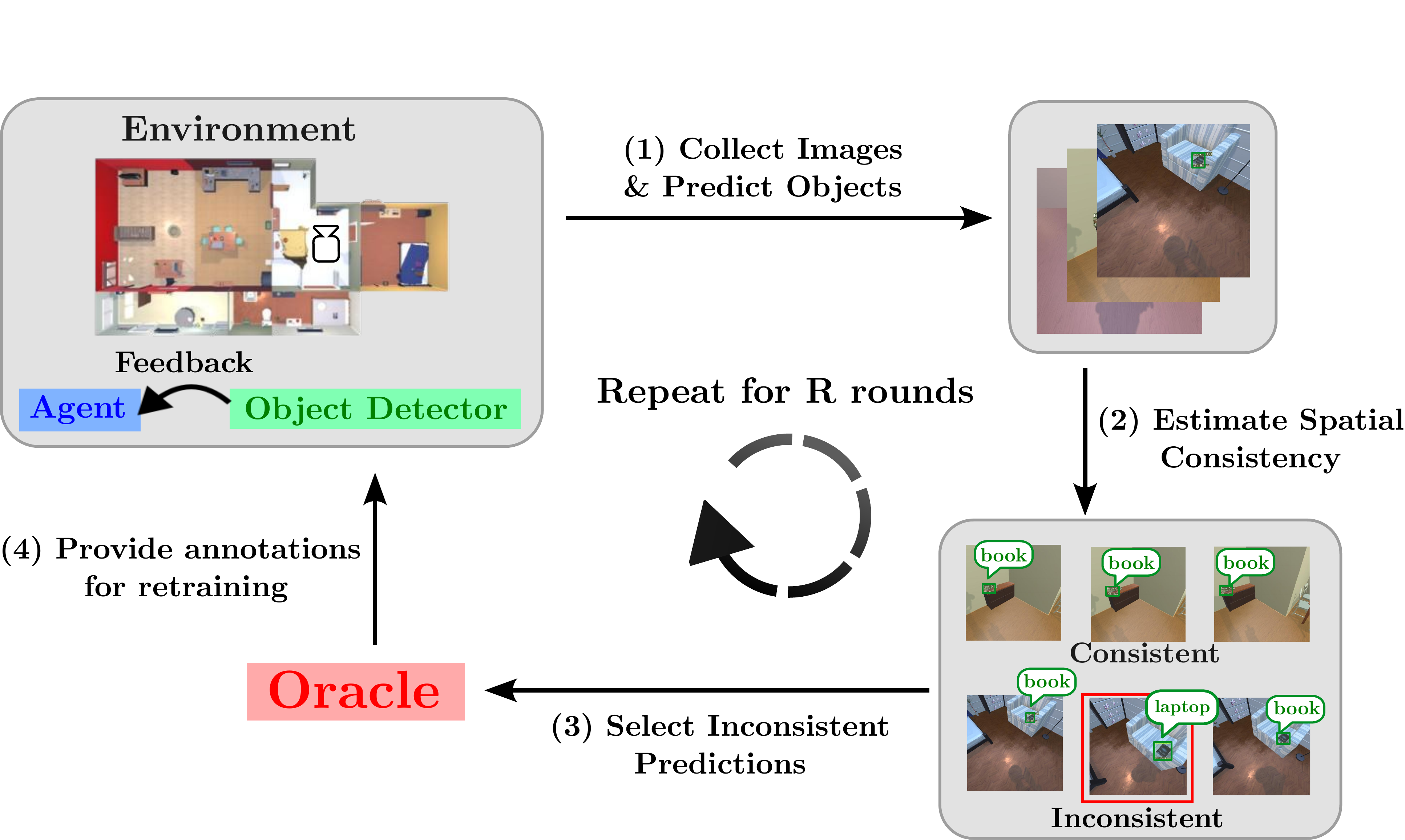}
    \caption{Method overview: (1) An agent explores the environment via planning, capturing images and detector predictions at each step; the detector's feedback encourages exploring error-prone states in future trajectories. (2) We estimate the spatial consistency of consecutive predictions and select a diverse set of inconsistent samples (3), which are annotated by an oracle (4) to improve the detector. The process repeats iteratively.}
\end{figure}

This paper investigates autonomous data collection and selection methods to adapt robotic vision models to a novel, previously unseen environment, with the help of an oracle annotator. We aim to address the following questions:
    \begin{itemize}
    \item How can a robot agent adapt its vision model to a new environment under limited navigation and annotation budgets?
    \item What scoring method should the agent use to determine which images are most informative to annotate, without relying on external supervision?
    \item What type of navigation strategy enables the agent to locate the hardest-to-detect objects in a scene?
\end{itemize}

In this work, our main contributions are as follows:
\begin{itemize}
    \item We formulate the problem of vision model adaptation as an embodied version of batch active learning, with limited budget for both navigation and annotation.
    \item We introduce a Prediction Discrepancy score that measures the disagreement between consecutive predictions, and use this score to guide both the navigation policy and the image sampling strategy to informative samples.
    \item We evaluate and compare our method using house layouts from the ProcTHOR dataset \cite{kolve_ai2-thor_2022, deitke_procthor_2022}.
    \item We further validate our approach in a real-world setup, where a robot autonomously collects images to improve its object detector using a semi-automatic process utilizing foundation models as an oracle.
\end{itemize}

\section{Related Work} \label{sec:related_work}

\textbf{Object detection} is a classic and widely studied problem in computer vision \cite{zhao_object_2019}. Two-stage methods \cite{girshick_rich_2014,ren_faster_2015} first generate region proposals, then refine and classify those regions, which tends to improve accuracy at the cost of speed. One-stage methods \cite{liu_ssd_2016,redmon_you_2016,redmon_yolov3_2018,jocher_ultralyticsyolov5_2022} skip the proposal step and directly predict object classes and bounding boxes in a single pass, enabling real-time performance and making them widely used in robotics applications. More recent transformer-based approaches like DETR \cite{carion_end--end_2020} introduce an end-to-end set prediction paradigm that avoids both region proposals and anchor-based predictions entirely. In this work, we propose a method that is agnostic to the detector architecture, illustrated on the YOLO family of real-time detectors.

\textbf{Active learning for object detection.}
Active learning techniques aim to minimize the annotation cost by selecting the most relevant samples to annotate given the model current capabilities. They typically combine difficulty estimation with diversity estimation to construct a set of diverse, informative samples for annotation. 
Instance difficulty can be estimated using prediction entropy \cite{joshi_multi-class_2009, ayers_query-based_2023}, posterior probabilities \cite{lewis_heterogeneous_1994}, or learned loss prediction \cite{yoo_learning_2019}.
Yang et al. \cite{yang_plug_2024} proposed a simple two-stage method, where the first stage selects difficult samples and the second stage performs diversity-based sampling. 
In this work, we adopt a similar two-stage strategy with the same diversity criterion, but introduce custom difficulty estimators tailored to our low-data setting with an embodied agent. 

\textbf{Spatial consistency as a supervision signal.}
Prediction consistency across multiple viewpoints of the same object, has proven to be an effective supervision signal for training vision models. Yu et al. \cite{yu_consistency-based_2022} use prediction inconsistency over multiple image augmentations to estimate instance difficulty in active learning for object detection. Chaplot et al. \cite{chaplot_semantic_2020} use spatial inconsistency as a reward signal for a navigation agent in an embodied active learning setting; however, their approach assumes that full agent trajectories are annotated by the oracle and therefore does not address restricted annotation budget. Chaplot et al. \cite{chaplot_seal_2021} aggregate multiple views collected by an agent to disambiguate object categories and generate stronger pseudo-labels, which are then propagated to all views to retrain an object detector. In contrast, we exploit spatio-temporal inconsistency primarily as an uncertainty signal to guide navigation and annotation under strict annotation constraints.

\textbf{Curiosity for efficient environment exploration.}
In embodied active learning, candidate samples are collected by an autonomous agent whose navigation is not guided by task-specific extrinsic rewards. 
A classical exploration algorithm is Frontier-Based Exploration (FBE) \cite{yamauchi_frontier-based_1997}, which consists of sampling next destinations on the frontier between explored and unexplored regions. 
Inspired by reinforcement learning \cite{Sutton1998} and curiosity \cite{oudeyer_what_2007} theory, researchers have introduced intrinsic reward signals to encourage agent exploration by rewarding model disagreement \cite{pathak_curiosity-driven_2017} or model prediction error \cite{burda_exploration_2018}. 
Marza et al. \cite{marza_autonerf_2023} implement an adaptive curious agent for mapping previously unknown scenes under a fixed time budget by training a PPO \cite{schulman_proximal_2017} policy to select waypoints. The collected images are then used for mesh reconstruction with a NeRF. 
Chaplot et al. \cite{chaplot_semantic_2020} similarly use spatial inconsistency as an intrinsic reward to train a PPO policy across many scenes. In this work, we leverage spatial inconsistency as an artificial curiosity signal for pure exploration, using it to steer the agent toward images that offer the highest potential for improving object detector performance.

\section{method}\label{sec:overall_method}

\subsection{Task Definition}
\label{subsec:task_definition}

We formulate visual model adaptation as an embodied variant of batch active learning, where a mobile robot has limited resources for both navigation (to collect candidate samples) and annotation (to label the most informative ones).

We follow the environment specification of \cite{zheng_towards_2022}. An agent state $s$ is defined by a 2D coordinate $(x, z)$ in a discrete grid, discrete pitch and yaw angles ($v$, $w$). At each time step, the agent can move forward, backward, left, or right by 25 cm; rotate left or right by 45°; look up or down by 30°.

At each timestep, the agent observes an RGB frame $I_s$, and the object detector $\mathcal{M}_\theta$ predicts all objects $o$ in the frame. Each prediction consists of a class $c$ in a fixed set of classes $C$ and a bounding box $(x,y,w,h)$. We denote by $O^\theta_s = \{o^\theta_{s,i}\}$ the set of predicted objects and by $Y_s = \{y_{s,i}\}$ the ground-truth objects for frame $I_s$.

\begin{figure*}[ht]
    \centering
    \includegraphics[width=0.71\textwidth]{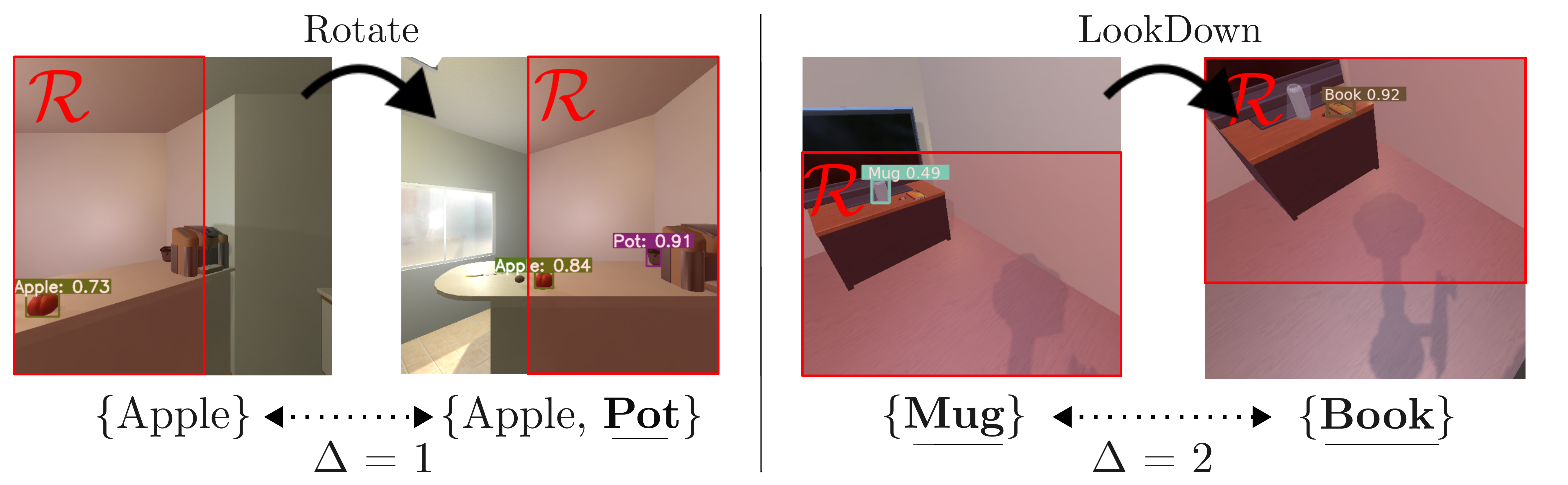}
    \vspace{-0.125cm}
    \caption{Examples of Prediction Discrepancy ($\Delta$) for two pairs of successive frames. The red area denotes the overlapping region $\mathcal{R}$ between the two successive frames. We match predicted objects of the same class across frames and count unmatched objects. (Left) Apple is matched but Pot is unmatched: $\Delta = 1$. (Right) Both Mug and Book are unmatched: $\Delta = 2$. The Prediction Discrepancy allows the agent to estimate model prediction errors without external supervision.}
    \label{fig:discrepency_reward}
    \vspace{-0.2cm}
\end{figure*}

\textbf{Data Collection Protocol.} We divide our adaptation method into rounds, each round consisting of a collection phase, selection phase, retraining phase, and an optional evaluation phase.

In the collection phase, the agent has $T$ timesteps to collect candidate samples, storing visited frames and predictions in a collection buffer $\mathcal{B}_c = \{(I_s, O^\theta_s)\}$, accumulated from round $0$ up to round $r$. 

In the selection phase, the agent selects a fixed number $A < T$ of the most relevant samples from $\mathcal{B}_c$ and requests annotations from the oracle. These samples are moved from $\mathcal{B}_c$ to the annotation buffer $\mathcal{B}_a = \{(I_s, Y_s)\}$.  

In the retraining phase, the agent retrains the object detector on $\mathcal{B}_a$, obtaining retrained model $\mathcal{M}_{\theta'}$, and updates predictions in $\mathcal{B}_c$ to $O^{\theta'}_s$. In the evaluation phase, $\mathcal{M}_{\theta'}$ is evaluated on a held-out set of positions and labels $\{(I_s, Y_s)\}$ from the same scene.

\subsection{Prediction Discrepancy}
\label{subsec:discrepancy}
We define \textit{Prediction Discrepancy}, a heuristic to rank candidate samples $(I_{s}, O^{\theta}_{s})$ based on the disagreement between model predictions in consecutive states. 

Formally, suppose the agent is at state $s$, performs action $a$, and reaches state $s'$. Let $\mathcal{R}(a)$ be the overlapping region between frames $I_s$ and $I_{s'}$, and $C^{\theta}_{s|\mathcal{R}(a)}$ the list of classes of objects from $O^{\theta}_s$ restricted to $\mathcal{R}(a)$. The prediction discrepancy is the size of the symmetric difference between $C^{\theta}_{s|\mathcal{R}(a)}$ and $C^{\theta}_{s'|\mathcal{R}(a)}$:
\begin{equation}
    \begin{split}
    \vspace{-0.2cm}
    \Delta(\theta,I_{s}, I_{s'},a) 
    = |C^{\theta}_{s|\mathcal{R}(a)} \cup C^{\theta}_{s'|\mathcal{R}(a)}| \\
    - |C^{\theta}_{s|\mathcal{R}(a)} \cap C^{\theta}_{s'|\mathcal{R}(a)}|
    \end{split}
    \label{equation:discrepency_reward_pair}
\end{equation}
treating both lists as multisets (duplicate classes are counted); in practice, we match elements of identical class across the two lists and count the remainder as unmatched. $\mathcal{R}(a)$ is estimated heuristically from the action alone, without image or depth information. We exclude unmatched elements from the computation if they are too small or near its border.

$\Delta = 0$ indicates no disagreement between the two frames, while $\Delta > 0$ means some objects are detected in one frame but missed or labeled differently in the other, revealing prediction errors without requiring external supervision. 

Note that this measure does not account for localization errors: two bounding boxes from different images may be matched but not necessarily geometrically align. Enforcing such localization alignment would require estimating objects position relative to the agent across the two frames, and thus a depth estimation module.

The discrepancy for a single frame $I_s$ is obtained by averaging over the pairs of visited neighbors, so a frame scores high if the model is likely to change its prediction upon taking an action from that position.

\subsection{Adaptive Navigation Agent}
\label{subsec:bo_agent}

We propose an online optimization algorithm for a curious agent operating under a limited navigation budget. To guide exploration, we use a path-aware Bayesian optimization (BO) strategy that balances optimistic exploration with maximization of the Prediction Discrepancy.

The approach relies on a path-aware BO framework \cite{santejudean_path-aware_2021,frazier_tutorial_2018}, using a Random Forest regressor \cite{breiman_random_2001} as a surrogate model to estimate $s \mapsto \Delta(\theta, I_s)$ from the visited states in the collection buffer $\mathcal{B}_{c} = \{(I_s, O^{\theta}s)\}$. Optimistic value estimates are obtained via an Upper Confidence Bound (UCB) \cite{ucb_paper}, defined as $\mu_{\text{UCB}}(s) = \mu(s) + 3\sigma(s)$, where $\mu(s)$ and $\sigma(s)$ denote the predictive mean and variance estimated by the forest.

Let $s_{\text{start}}$ be the starting position, and let $\pi(s_{\text{start}}, s_{\text{dest}})$ denote the shortest path between $s_{\text{start}}$ and destination $s_{\text{dest}}$. We define the acquisition function as an optimistic estimate of the average Prediction Discrepancy along this path:

\begin{equation}
\alpha(s_{\text{dest}}) = \frac{1}{|\pi(s_{\text{start}}, s_{\text{dest}})|} \sum_{s' \in \pi(s_{\text{start}}, s_{\text{dest}})} \mu_{\text{UCB}}(s').
\label{equation:bo_acquisition}
\end{equation}

The next destination is the candidate position that maximizes $\alpha(s_{\text{dest}})$ among 50 randomly sampled positions; the agent follows the trajectory to the target before repeating the process.

The shortest paths $\pi(s_{\text{start}}, s_{\text{dest}})$ are computed by weighted A*. The graph edge weights---a cost in $[0, 1]$ to be minimized along the planned path---are designed to encourage movement toward positions with high optimistic estimates $\mu_{\text{UCB}}(s)$ while discouraging revisiting locations:

\begin{equation}
w(s) =
\begin{cases}
 \frac{1 - \bar{\mu}_{\text{UCB}}(s)}{2} & \text{if } s \notin \mathcal{V}, \\
1 & \text{otherwise},
\end{cases}
\end{equation}

where $\bar{\mu}_{\text{UCB}}$ denotes $\mu_{\text{UCB}}$ normalized to $[0, 1]$, and $\mathcal{V}$ represents the set of previously visited positions. This weighting guides the agent toward positions that are considered promising based on its current model, while avoiding revisiting already explored areas. Before each acquisition, the surrogate model is recomputed. After each model retraining, the predictions from the collection buffer are updated and the surrogate model is recomputed.

\subsection{Data Selection} 
\label{subsec:data_selection}
At each round, the agent selects a diverse subset of $A$ samples from $\mathcal{B}_c$ that is most informative for training, from samples collected since the start of the run.

We follow the two-stage active learning approach proposed by PPAL \cite{yang_plug_2024} to combine difficulty-sampling and diversity-sampling. In the first stage, we select the $3 \times A$ samples with the highest Prediction Discrepancy $\Delta(\theta, I_s)$. In the second stage, we cluster them into $A$ groups and select each cluster centroid for annotation.
To estimate image distances for the clustering algorithm, PPAL introduces Class Conditioned Matching Similarity (CCMS): two frames $I_s$ and $I_{s'}$ are similar if they contain objects $O_s^{\theta}$ and $O_{s'}^{\theta}$ of the same classes with aligned features. We keep CCMS unchanged, but use features from a frozen ResNet encoder instead of the YOLO object detector $\mathcal{M}_\theta$ for simplicity. We modify the second stage of PPAL to ensure sampled images differ from the annotated set $\mathcal{B}_a$. We combine the $n$ annotated samples with the $3 \times A$ candidates before clustering, fit $A + n$ clusters, and discard clusters corresponding to annotated frames, effectively removing candidates too close to previous annotations. 

The selected data is annotated by the oracle, and it is used to retrain the object detector. We accumulate the annotated data from all previous rounds and load the initial weights before each retraining. Algorithm~\ref{alg:eal} provides an overview of how planning, data selection and model updating interact with each other.   

\algnewcommand{\Initialize}[1]{%
  \State \textbf{Initialize:}
  \Statex \hspace*{\algorithmicindent}\parbox[t]{.8\linewidth}{\raggedright #1}
}

\begin{algorithm}
    \small
    \caption{Embodied Active Learning under limited annotation and navigation budget}
    \begin{algorithmic}
    \Require{Number of rounds R, number of annotations per round A, number of timesteps per round T, initial model weights $\theta_0$}
    \State Initialize collection buffer $\mathcal{B}_c$, annotation buffer $\mathcal{B}_a$
    \State Initialize object detector $\mathcal{M}_{\theta}$
    \State Initialize current trajectory  $\pi =  \emptyset$
    \For{round r = 1 to R}
        \For{t = 1 to T}
        \If{$\pi =  \emptyset$} 
            \State Select a destination $s_{\text{dest}}$ using BO (See \ref{subsec:bo_agent})
            \State Set $\pi$ to the shortest path to $s_{\text{dest}}$
        \EndIf
        \State Execute next action $a$ from $\pi$ and remove it from $\pi$
        \State Observe frame $I_s$ and predict objects $O^\theta_s \leftarrow \mathcal{M}_\theta(I_s)$
        \State Add sample to buffer $\mathcal{B}_c \leftarrow \mathcal{B}_c \cup \{(I_s, O^\theta_s)\}$
        \State Update the agent's surrogate model (See \ref{subsec:bo_agent})
      \EndFor
    \State Select A samples $\{(I_{s_i},O^\theta_{s_i})\}$ from $\mathcal{B}_c$  using the scoring method $\Delta(\theta,I_s)$  (See \ref{subsec:data_selection})
    \State Query oracle for annotations $\{(I_{s_i}, Y_{s_i})\}$
    \State Add annotations to buffer $\mathcal{B}_a \leftarrow \mathcal{B}_a \cup \{(I_{s_i}, Y_{s_i})\}$
    \State $\theta' \leftarrow \text{RETRAIN}(\theta_0,\mathcal{B}_a)$
    \State Update model $\mathcal{M}_\theta \leftarrow \mathcal{M}_{\theta'}$ 
    \State Update samples $(I_s,O^{\theta}_s)$  from $\mathcal{B}_c$ with new predictions $O^{\theta'}_s$ 
    \State Optional: Evaluate model on an evaluation set $\{(I_s, Y_s)\}$

    \EndFor
  \end{algorithmic}
    \label{alg:eal}
\end{algorithm}
\vspace{-0.35cm}

\vspace{-0.2cm}
\section{Simulation Experiments}
In this section, we describe our experiments in the simulated environment. We propose to compare different \textbf{scoring methods} (serves to rank images and guide agents) and different \textbf{navigation agents} (strategies to explore the scene).

\begin{figure*}[ht]
    \centering
    \includegraphics[width=0.95\linewidth]{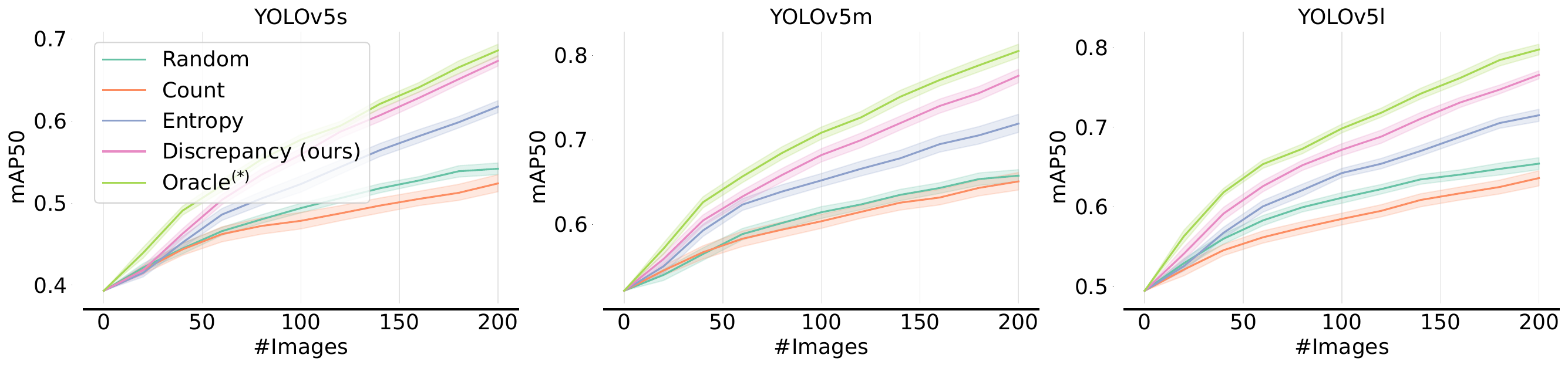}

    \vspace{-0.2cm}
    \caption{Aggregate Performance of different scoring methods, across different model sizes (left: small, middle: medium, right: large). Each method is evaluated with the same annotation and navigation budget, and use the Adaptive navigation agent. (*): the \textit{Oracle} scoring method uses ground-truth labels to select images containing classes with low accuracy. The relative performance of the scoring methods is consistent across model sizes.
    }
    \label{fig:xp_scores}
    \vspace{-0.2cm}
\end{figure*}

\subsection{Implementation Details}
\textbf{Environment.} We run our experiments using 90 house layouts from the ProcTHOR \cite{deitke_procthor_2022} dataset. We filter out object classes which are not pickable and those that are present in less than 20 scenes. Each scene has at least 2 rooms and between 12 and 40 object classes represented.
We use a $640\times640$ image size, and we filter out objects that occupy less than $24\times 24$ pixels of the images. The image annotations are provided by the simulator API. The relative performance of the methods is consistent across model sizes.

\textbf{Object Detector.} We run our experiments with the single-stage object detector YOLOv5 \cite{jocher_ultralyticsyolov5_2022}, and experiment with three model sizes S, M, and L, from smallest to largest. This model has open code, data and weights, with the original weights trained on the COCO dataset. We pretrain the model using 100 images randomly collected in each of the first 64 house scenes, and we run the model adaptation separately and independently on the 10 last scenes.  We keep original hyperparameters, except for the adaptation learning rate, set to $0.02$, the pretraining learning rate, set to $0.002$ and the epochs set to $50$ for both pretraining and adaptation. Original hyperparameters include data augmentations (mosaic, flip, and scale). During the collection phase, the object detector uses a confidence threshold set to $0.2$.

\textbf{Budget.} We run our collection method for $R = 10$ rounds, each round with $T=400$ navigation steps and $A = 20$ annotated samples.

\subsection{Baselines}
\textbf{Baseline Scoring Methods.}
We investigate alternatives to the Prediction Discrepancy.

\textbf{Count} estimates the image difficulty by counting the number of predicted objects: $\text{Count}(\theta,I_s) = |O^{\theta}_s|$.

\textbf{Entropy} estimates the image difficulty by summing the class-wise prediction entropy over all the predicted objects. This is a very common choice in active learning methods \cite{joshi_multi-class_2009, yang_plug_2024} and hard-image retrieval \cite{ayers_query-based_2023}:

\begin{equation}
    \mathcal{H}(\theta,I_s) = - \sum_{o \in O^{\theta}_s}  \sum_{c \in C} p_{\theta}(c(o)=c) \text{log}(p_{\theta}(c(o)=c))
    \label{equation:total_entropy}
\end{equation}

where $p_{\theta}(c(o)=c)$ refers to the probability that the predicted object $o$ belongs to class c.

\textit{Oracle} uses the ground-truth labels $Y_s$ to estimate the difficulty of the image based on the sum of the model accuracy gap on the corresponding classes:

\begin{equation}
    \Omega(\theta,I_s,Y_s) = \sum_{o \in Y_s}  [1 - \text{AP}_\theta(c(o))] 
    \label{equation:oracle}
\end{equation}
where $\text{AP}_\theta(c(o))$ refers to the Average Precision on the class of the object $o$. Oracle is a strong scoring method because it has an important sampling preference for the objects that currently have low performance. However, the Oracle scoring method requires to have access to the ground-truth objects $Y_s$ which is not available in practice. All of these criterion can be implemented as a feedback signal for the Adaptive navigation agent. We compare their performance in Figure \ref{fig:xp_scores}.

\textbf{Baseline Agents.} We compare the Adaptive navigation agent against three baseline agents. These three navigation strategies do not attempt to maximize the Prediction Discrepancy. We use Frontier-Based Exploration \cite{yamauchi_frontier-based_1997} because it is commonly used as a baseline in the robotics literature \cite{marza_autonerf_2023}, and we use a cone-of-vision with 90° angle and 1m depth. We modify the navigation strategy proposed for the pretraining in \cite{zheng_towards_2022}, by defining the \textit{Sweep} agent to be an agent that randomly samples a destination, walks to reach it, and sweeps all the possible angles. We also implement a simple \textit{Random} walk policy.
Because of the limited navigation budget, we do not propose an end-to-end RL baseline. 

\subsection{Evaluation set and Metrics}
We use a static evaluation set independent of the tested method. Each set contains 1,000 annotated images generated from random starting positions to locations where a queried object is visible. Every object class is queried at least once, ensuring full class coverage within the scene. Performance is evaluated on the same scene used for adaptation.

We evaluate the object detector using mean Average Precision (mAP50), a commonly used metric in object detection. It is defined as the mean over classes of the area under the Precision–Recall curve, computed at an IoU threshold of $0.5$.

\subsection{Results}
Each experiment is repeated independently on 10 scenes, each evaluated with 10 different random seeds. We follow the recommendations from Agarwal et al. \cite{agarwal2021deep} to aggregate metrics over all seeds and scenes using bootstrap confidence intervals and Interquartile Mean (IQM).

\textbf{Comparison of the scoring methods.}
Figure \ref{fig:xp_scores} presents a comparison of the performance in the simulated environment obtained with Prediction Discrepancy and the baseline scoring strategies: Entropy, Count, Random, and Oracle. The results show that Prediction Discrepancy consistently outperforms the standard baselines (Entropy, Count, and Random), highlighting its ability to identify more informative samples for training.

In contrast, the Oracle scoring method achieves the highest performance. This is expected, as it relies on ground-truth annotations to deliberately select samples containing difficult classes. While this method is not applicable in practice, it confirms that the model progress is driven by improvement on the most difficult classes.

\begin{figure}[t]
    \centering
    \vspace{-0.1cm}
    \includegraphics[width=0.485\textwidth]{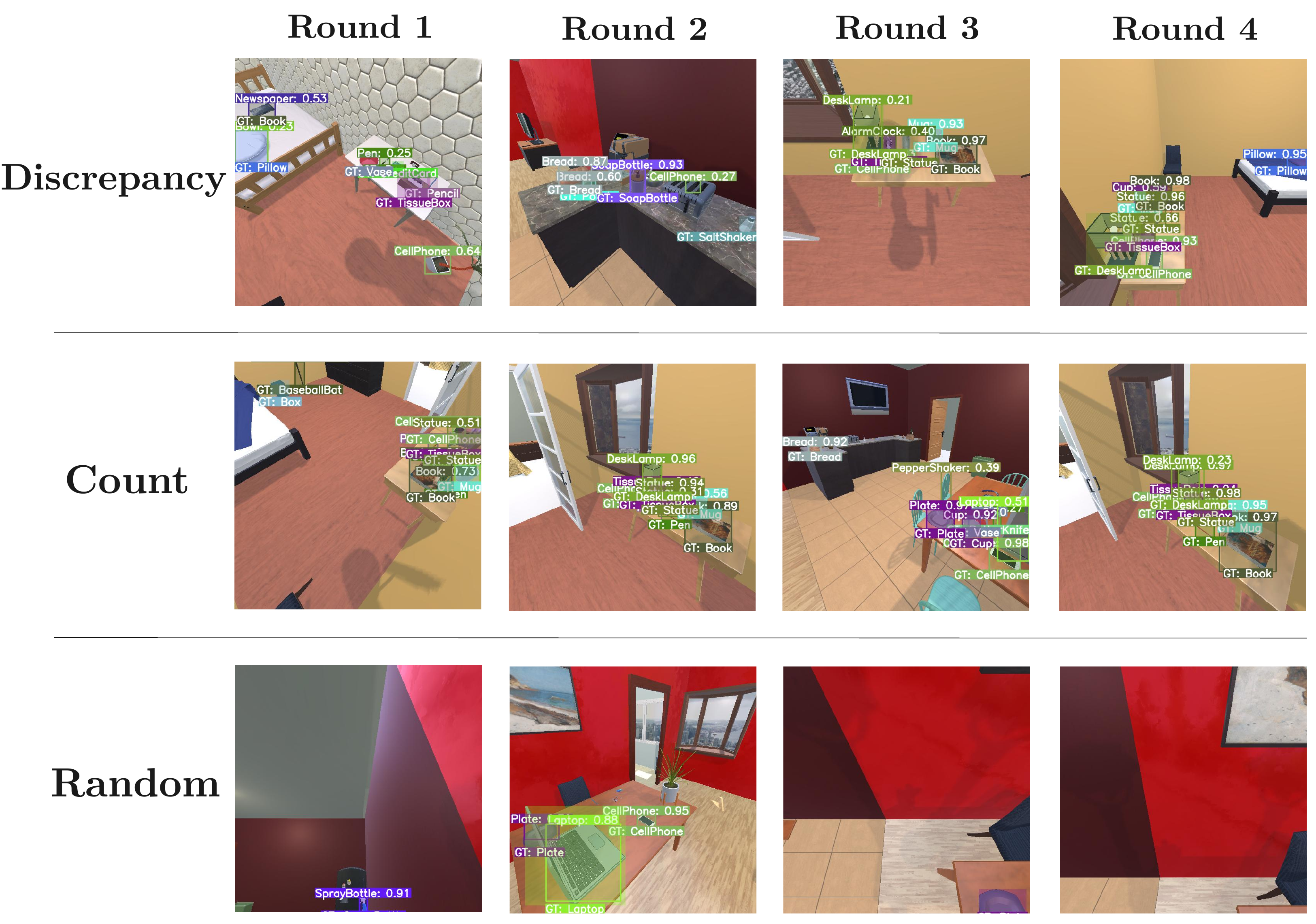}
    \vspace{-0.3cm}
    \caption{Collected data across 3 scoring methods. We visualize the first sample selected in each of the first four rounds for 3 different scoring methods. Count focuses only on cluttered images, Discrepancy adapts to the model’s weaknesses, and Random selection often yields irrelevant samples.
    }
    \vspace{-0.1cm}
    \label{fig:xp_collection}
\end{figure}

\textbf{Annotated Data Characteristics for different scoring methods.}
 We introduce three additional metrics to characterize the samples selected by each scoring method:
 
\begin{itemize}
    \item Object Annotations: average number of ground-truth objects in the samples annotated by the oracle per round. 
    \item Class Distribution Entropy: the entropy of the class distribution of the annotation buffer at the end of the experiment. Higher entropy means that the samples cover more evenly the classes.
    \item $F1_{\text{annot}}$: current model $F1$ score on the annotated samples. Lower F1 means that the selected samples are more challenging for the model.
\end{itemize}


\begin{table*}[ht]
    \normalsize
    \centering
    \caption{Annotated Data Characteristics for different scoring methods in AI2-THOR and Real-world environments. Object Annotations is the average number of ground-truth objects in the samples annotated by the oracle. Class Distribution Entropy is the entropy of the class distribution of the annotation buffer at the end of the experiment. $F1_{\text{annot}}$ is the model $F1$ score on annotated samples, averaged over the first 3 rounds. (*): the \textit{Oracle} scoring method uses ground-truth labels to select images containing classes with low accuracy.}
    \label{tbl:uncertainty}
    \begin{tabular}{l|ccc|ccc}
    \hline
    & \multicolumn{3}{c|}{\textbf{AI2-THOR}} & \multicolumn{3}{c}{\textbf{Real-World}} \\
    \cline{2-7}
    Scoring Method 
    & \#Object Annot. 
    & \small{Cls. Entropy}$(\uparrow)$ 
    & $F1_{\text{annot}}$$(\downarrow)$
    & \#Object Annot.
    & \small{Cls. Entropy}$(\uparrow)$ 
    & $F1_{\text{annot}}$$(\downarrow)$ \\
    \hline
    Random      
        & $14.33\pm0.84$    
        & $3.78\pm0.08$   
        & $77.78\pm1.87$
        & $63.47\pm8.47$  
        & $2.33\pm0.12$  
        & $46.08\pm4.60$ \\

    Count       
        & $103.26\pm4.50$   
        & $3.38\pm0.15$   
        & $83.29\pm2.01$
        & $184.07\pm4.22$ 
        & $2.08\pm0.05$  
        & $47.11\pm1.21$ \\

    Entropy     
        & $85.57\pm2.39$    
        & $3.68\pm0.10$   
        & $74.41\pm2.08$
        & $163.93\pm7.76$ 
        & $2.14\pm0.05$  
        & $44.62\pm0.83$ \\

    Discrepancy (ours)
        & $52.56\pm1.51$    
        & $3.96\pm0.05$   
        & $67.26\pm1.66$
        & $75.56\pm6.11$  
        & $2.28\pm0.07$  
        & $39.63\pm1.99$ \\

    $\text{Oracle}^{(*)}$ 
        & $76.22\pm2.75$    
        & $3.97\pm0.07$   
        & $68.36\pm1.90$
        & --- 
        & --- 
        & --- \\

    \hline
    \end{tabular}
\end{table*}
Table \ref{tbl:uncertainty} shows that different scoring methods yield different values for these metrics. Prediction Discrepancy and Oracle give the most difficult samples according to the average F1, while the other methods sample easier data. These two methods also achieve the highest Class Distribution Entropies, indicating more balanced and diverse class sampling, consistent with their stronger overall performance. Prediction Discrepancy also results in a lower number of Object Annotations. Examples of samples collected with the different scoring methods are illustrated in Figure \ref{fig:xp_collection}.

\begin{figure}[t]
    \centering
    \includegraphics[height=0.1625\textwidth]{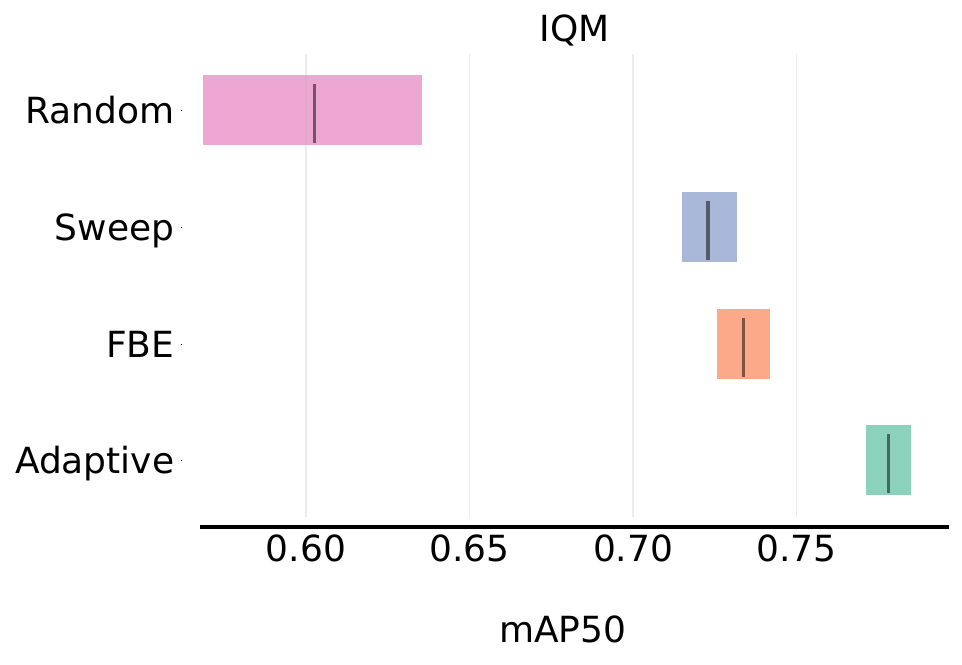}
    \includegraphics[height=0.1625\textwidth]{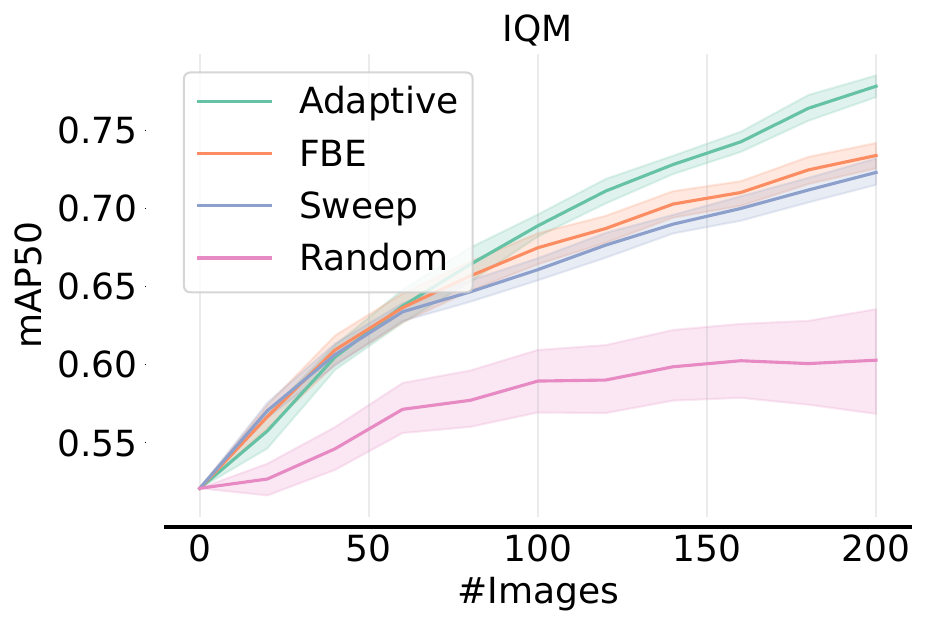}
    \caption{Aggregate Performance of different navigation methods. Each method is evaluated with the same annotation and navigation budget, and use the Discrepancy scoring method.}
    \label{fig:xp_navigation}
    \vspace{-0.2cm}
\end{figure}

\textbf{Comparison of the navigation methods.} In Figure \ref{fig:xp_navigation}, we compare our Adaptive navigation agent with the baseline agents in the simulated environment. On the long run, the Adaptive navigation agent yields a better learning progress compared to the Frontier-Based Exploration (FBE), Sweep, and Random walk baselines. This proves that guiding the agent with the Prediction Discrepancy is an important component of our approach. 

\section{Real-World Experiments}
We demonstrate our approach in a real-world scene with Boston Dynamics Spot\textsuperscript{\textregistered}~\cite{spot} as the embodied agent. We begin by describing our environment, data collection process and budget, the trajectory selection approach, followed by the data labeling method. Finally, we describe our results.

\subsection{Implementation Details}
\begin{figure}
    \centering
    \includegraphics[width=\linewidth]{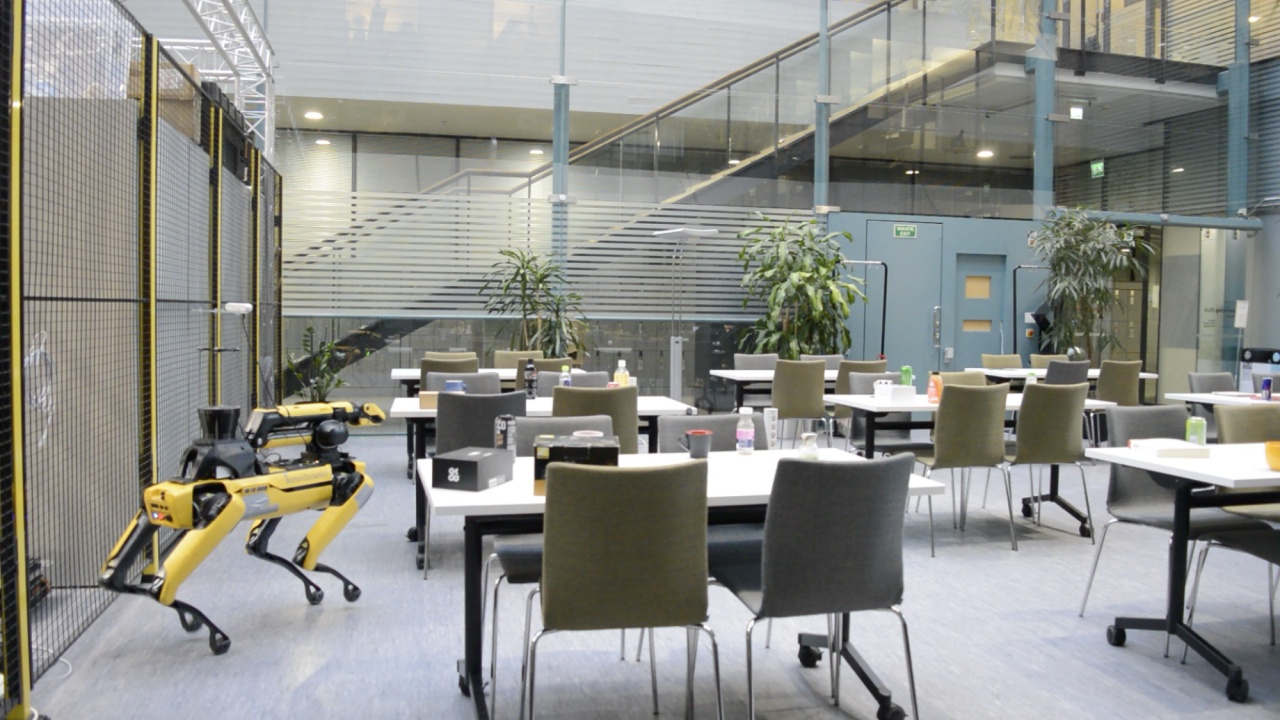}
    \caption{Indoor experiment environment.}
    \vspace{-0.15cm}
    \label{fig:indoor_scene}
\end{figure}

\begin{figure}[h]
    \centering
    \includegraphics[width=\linewidth]{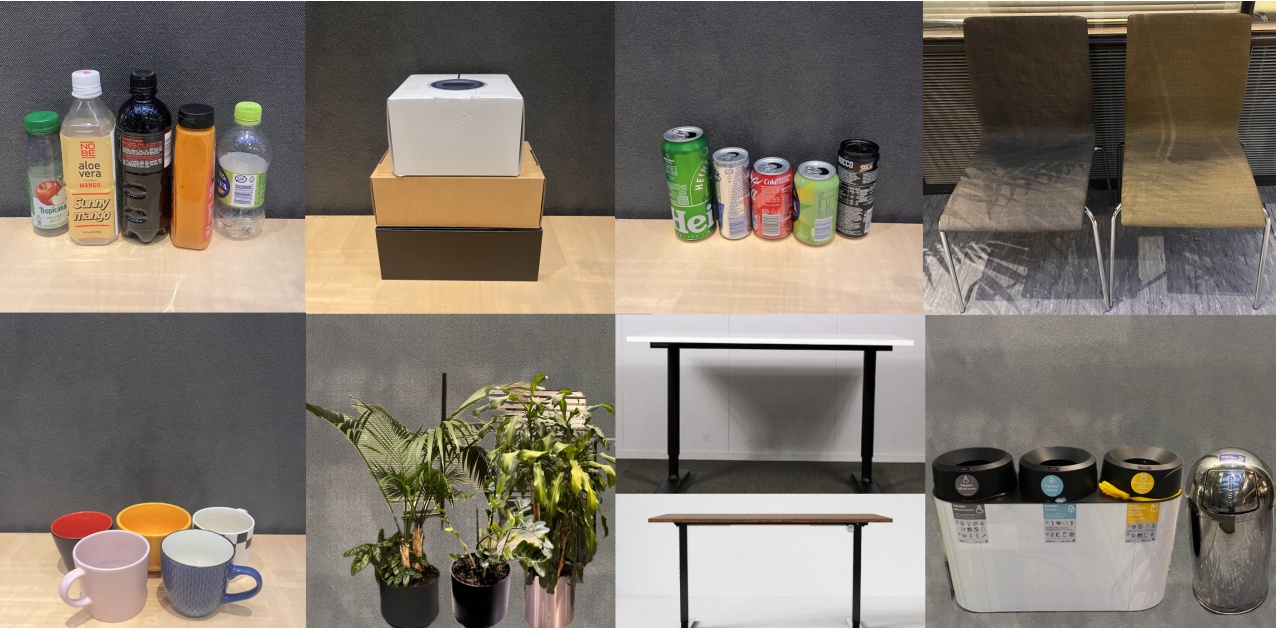}
    \caption{Objects used for real robot experiments: bottle, box, can, chair, cup, plant, table, and trashcan.}
    \vspace{-0.15cm}
    \label{fig:object_classes}
\end{figure}

\textbf{Environment.} 
Our environment consists of an indoor scene as shown in Figure \ref{fig:indoor_scene}, with objects from eight categories and 2–5 physical variations per class (Figure~\ref{fig:object_classes}). 

The robot state $s$ is represented by its planar position $(x,y)$, yaw angle, and body height. We discretize the state space using step sizes of $0.15$ meters for position, $10^\circ$ for yaw, and three discrete bins for body height. The robot can execute the following actions: forward, backward, left, right, rotate clockwise, rotate counterclockwise, sit, stand, and crouch. We use a $640\times640$ image size.

\textbf{Data Collection.}
We first collect a pretraining dataset of 100 images in the scene with objects arranged in different configurations. After modifying the scene, we store exploration trajectories by navigating the robot between predefined waypoints. To increase viewpoint diversity, we inject random rotations, crouching, and sitting actions along the planned routes. The recorded data are downsampled to 3 Hz to ensure moderate overlap between successive frames. We store 50 trajectories of $T=200$ navigation steps each. Additionally, we store a held-out set of 200 images and labels $\{(I_s, Y_s)\}$ for quantitative comparison of the scoring methods. We run our collection method for $R = 5$ rounds, each round with 10 trajectories of $T=200$ navigation steps and $A = 20$ annotated samples. 

We reuse the Adaptive navigation agent (Section~\ref{subsec:bo_agent}), assigning each stored trajectory a score based on the sum of the acquisition values along its path. The highest-scoring trajectory is then selected and appended to the collection buffer $\mathcal{B}_c$ for the next stage.

\textbf{Object Detector.} We use YOLOv5 with M and L model sizes. We keep the same hyperparameters as simulation experiments, except for the pretraining learning rate, set to $0.05$ and the epochs set to $100$ for pretraining.

\textbf{Overlapping region.} Unlike the heuristic computation used in the simulated environment to estimate the overlapping region between two frames, we use a homography-based method for the real-world environment. We detect feature correspondences between successive frames and apply RANSAC to robustly compute the homography matrix, from which the overlapping rectangle is derived. 

\textbf{Vision Foundation Model as Oracle:} Manually annotating real-world images is time-consuming, so we explore a semi-automated labeling pipeline using foundation models. We use OWLv2 \cite{minderer2023scaling}, an open-vocabulary vision–language detector that localizes objects from arbitrary text queries. We manually define the object classes and provide them as input queries to OWLv2. Labeling a single image takes \textbf{$\sim$4.5 seconds} on an RTX 5090 GPU — a bottleneck that further motivates lightweight detectors at collection time. The generated labels are then corrected by a human.

\subsection{Results}
Each experiment is repeated for 15 seeds. Results are aggregated over all seeds using bootstrap confidence intervals and Interquartile Mean (IQM).

\begin{figure}[h]
    \centering
    \includegraphics[height=0.1291\textheight]{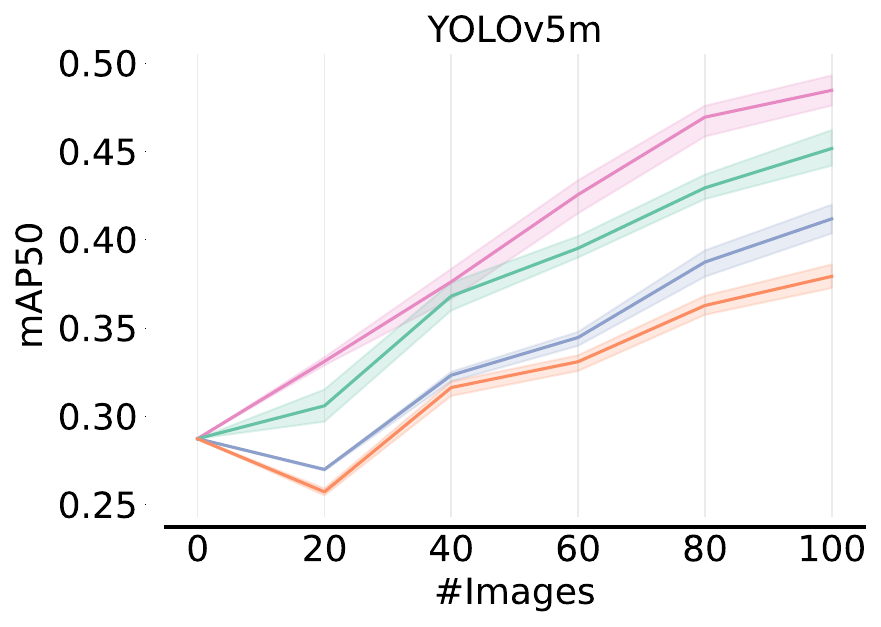}
    \includegraphics[height=0.1291\textheight]{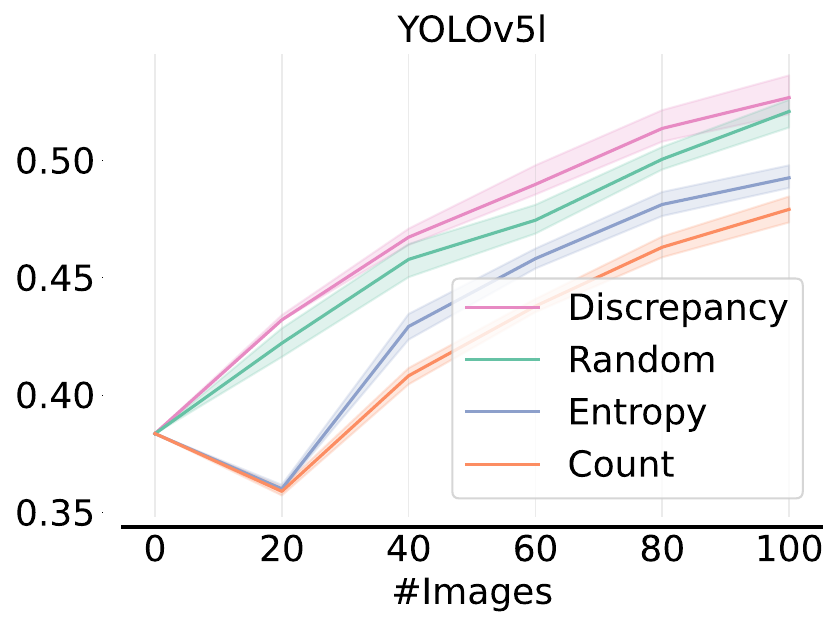}
    \caption{Aggregate Performance of different scoring methods across different model sizes (left: medium, right: large). Each method is evaluated with the same collection and budget. }
    \label{fig:results_rw}
    \vspace{-0.1cm}
\end{figure}

\textbf{Comparison of the scoring methods.}
Figure~\ref{fig:results_rw} compares the performance achieved by Prediction Discrepancy against the baseline scoring strategies: Entropy, Count, and Random. We exclude the Oracle method from this evaluation because access to ground-truth labels is unrealistic in real-world deployments. The results demonstrate that Prediction Discrepancy consistently outperforms the baselines, confirming its effectiveness in selecting more informative samples.

The performance gap between Random and Prediction Discrepancy is smaller than in simulation. We attribute this to the simplified single-room experimental setup and smaller budget, which limits environmental variability. Nevertheless, even in smaller and less complex environments, Prediction Discrepancy maintains an advantage.

Table \ref{tbl:uncertainty} shows the quantitative comparison between different scoring methods for the real-world scene. Consistent with the simulated environment, Prediction Discrepancy prioritizes difficult samples while using fewer object annotations than Count and Entropy-based scoring methods.

\section{Conclusion}
We presented an embodied active learning approach that adapts object detectors to new environments under both navigation and annotation budget constraints, improving downstream performance while reducing annotation cost. For future work, enabling the agent to explicitly estimate class difficulty and learning progress could help prioritize difficult or underrepresented classes and yield further gains. Additionally, we plan to explore models without domain-specific pretraining, investigate region-level annotation and self-supervised adaptation methods to reduce reliance on an oracle annotator.


\section*{Authors Contributions}
This work started during Mohamed Yassine Kabouri's internship at Inria, where he conducted initial research on Prediction Discrepancy optimization for image collection. Hadrien Crassous continued the work, transitioning it to an active learning formulation, developed the image sampling method and the Adaptive navigation agent, and conducted the simulation experiments. Minahil Raza designed and conducted all the real-world experiments. Joni Pajarinen and Riad Akrour advised the work.

\section*{Acknowledgements}
This research was supported by Inria via AEx AuDaCiTi and by the European Union's Horizon Europe research and innovation program under grant agreement No 101189836 (XSCAVE).
Experiments presented in this paper were carried out using the Grid'5000 testbed, supported by a scientific interest group hosted by Inria and including CNRS, RENATER and several Universities as well as other organizations. The authors acknowledge the use of the MIDAS infrastructure of the Aalto School of Electrical Engineering for the robot experiments.

{\small
\bibliographystyle{ieeetr}
\bibliography{references.bib}

@incollection{lewis_heterogeneous_1994,
	title = {Heterogeneous {Uncertainty} {Sampling} for {Supervised} {Learning}},
	copyright = {https://www.elsevier.com/tdm/userlicense/1.0/},
	isbn = {978-1-55860-335-6},
	url = {https://linkinghub.elsevier.com/retrieve/pii/B978155860335650026X},
	doi = {10.1016/B978-1-55860-335-6.50026-X},
	language = {en},
	urldate = {2026-02-12},
	booktitle = {Machine {Learning} {Proceedings} 1994},
	publisher = {Elsevier},
	author = {Lewis, David D. and Catlett, Jason},
	year = {1994},
}

@book{Sutton1998,
  author = {Sutton, Richard S. and Barto, Andrew G.},
  edition = {Second},
  publisher = {The MIT Press},
  title = {Reinforcement Learning: An Introduction},
  url = {http://incompleteideas.net/book/the-book-2nd.html},
  year = {2018 }
}

@inproceedings{yamauchi_frontier-based_1997,
	title = {A frontier-based approach for autonomous exploration},
	url = {https://ieeexplore.ieee.org/document/613851/},
	doi = {10.1109/CIRA.1997.613851},
	urldate = {2026-02-12},
	booktitle = {Proceedings 1997 {IEEE} {International} {Symposium} on {Computational} {Intelligence} in {Robotics} and {Automation} {CIRA}'97. '{Towards} {New} {Computational} {Principles} for {Robotics} and {Automation}'},
	author = {Yamauchi, B.},
	year = {1997},
}

@article{breiman_random_2001,
	title = {Random {Forests}},
	volume = {45},
	issn = {1573-0565},
	url = {https://doi.org/10.1023/A:1010933404324},
	doi = {10.1023/A:1010933404324},
	language = {en},
	number = {1},
	urldate = {2026-02-16},
	journal = {Machine Learning},
	author = {Breiman, Leo},
	month = oct,
	year = {2001},
	pages = {5--32},
}

@article{oudeyer_what_2007,
	title = {What is intrinsic motivation? {A} typology of computational approaches},
	volume = {1},
	issn = {1662-5218},
	shorttitle = {What is intrinsic motivation?},
	url = {https://www.frontiersin.org/journals/neurorobotics/articles/10.3389/neuro.12.006.2007/full},
	doi = {10.3389/neuro.12.006.2007},
	language = {English},
	urldate = {2026-02-12},
	journal = {Frontiers in Neurorobotics},
	publisher = {Frontiers},
	author = {Oudeyer, Pierre-Yves and Kaplan, Frederic},
	month = nov,
	year = {2007},
}

@inproceedings{joshi_multi-class_2009,
	title = {Multi-class active learning for image classification},
	issn = {1063-6919},
	url = {https://ieeexplore.ieee.org/document/5206627/},
	doi = {10.1109/CVPR.2009.5206627},
	urldate = {2026-02-12},
	booktitle = {CVPR},
	author = {Joshi, Ajay J. and Porikli, Fatih and Papanikolopoulos, Nikolaos},
	year = {2009},
	note = {ISSN: 1063-6919},
}

@article{everingham_pascal_2010,
	title = {The {Pascal} {Visual} {Object} {Classes} ({VOC}) {Challenge}},
	volume = {88},
	issn = {1573-1405},
	url = {https://doi.org/10.1007/s11263-009-0275-4},
	doi = {10.1007/s11263-009-0275-4},
	language = {en},
	number = {2},
	urldate = {2026-02-12},
	journal = {International Journal of Computer Vision},
	author = {Everingham, Mark and Van Gool, Luc and Williams, Christopher K. I. and Winn, John and Zisserman, Andrew},
	month = jun,
	year = {2010},
	pages = {303--338},
}

@inproceedings{girshick_rich_2014,
	author = {Girshick, Ross and Donahue, Jeff and Darrell, Trevor and Malik, Jitendra},
	title = {Rich Feature Hierarchies for Accurate Object Detection and Semantic Segmentation},
	booktitle = {CVPR},
	year = {2014}
}

@inproceedings{lin_microsoft_2015,
	title = {Microsoft {COCO}: {Common} {Objects} in {Context}},
	isbn = {978-3-319-10602-1},
	shorttitle = {Microsoft {COCO}},
	doi = {10.1007/978-3-319-10602-1_48},
	language = {en},
	booktitle = {ECCV},
	author = {Lin, Tsung-Yi and Maire, Michael and Belongie, Serge and Hays, James and Perona, Pietro and Ramanan, Deva and Dollár, Piotr and Zitnick, C. Lawrence},
	year = {2014},
}

@inproceedings{liu_ssd_2016,
	title = {{SSD}: {Single} {Shot} {MultiBox} {Detector}},
	isbn = {978-3-319-46448-0},
	shorttitle = {{SSD}},
	doi = {10.1007/978-3-319-46448-0_2},
	language = {en},
	booktitle = {ECCV},
	author = {Liu, Wei and Anguelov, Dragomir and Erhan, Dumitru and Szegedy, Christian and Reed, Scott and Fu, Cheng-Yang and Berg, Alexander C.},
	year = {2016},
}

@inproceedings{ren_faster_2015,
	title = {Faster {R}-{CNN}: {Towards} {Real}-{Time} {Object} {Detection} with {Region} {Proposal} {Networks}},
	shorttitle = {Faster {R}-{CNN}},
	url = {https://proceedings.neurips.cc/paper/2015/hash/14bfa6bb14875e45bba028a21ed38046-Abstract.html},
	urldate = {2026-02-16},
	booktitle = {Advances in {Neural} {Information} {Processing} {Systems}},
	publisher = {Curran Associates, Inc.},
	author = {Ren, Shaoqing and He, Kaiming and Girshick, Ross and Sun, Jian},
	year = {2015},
}

@inproceedings{redmon_you_2016,
	author = {Redmon, Joseph and Divvala, Santosh and Girshick, Ross and Farhadi, Ali},
	title = {You Only Look Once: Unified, Real-Time Object Detection},
	booktitle = {CVPR},
	year = {2016}
}

@misc{schulman_proximal_2017,
	title = {Proximal {Policy} {Optimization} {Algorithms}},
	url = {http://arxiv.org/abs/1707.06347},
	doi = {10.48550/arXiv.1707.06347},
	urldate = {2026-02-12},
	publisher = {arXiv},
	author = {Schulman, John and Wolski, Filip and Dhariwal, Prafulla and Radford, Alec and Klimov, Oleg},
	year = {2017},
	note = {arXiv:1707.06347 [cs]},
}

@inproceedings{maiettini_interactive_2017,
	title = {Interactive data collection for deep learning object detectors on humanoid robots},
	issn = {2164-0580},
	url = {https://ieeexplore.ieee.org/document/8246973/},
	doi = {10.1109/HUMANOIDS.2017.8246973},
	urldate = {2026-02-12},
	booktitle = {2017 {IEEE}-{RAS} 17th {International} {Conference} on {Humanoid} {Robotics} ({Humanoids})},
	author = {Maiettini, Elisa and Pasquale, Giulia and Rosasco, Lorenzo and Natale, Lorenzo},
	year = {2017},
	note = {ISSN: 2164-0580},
}

@misc{burda_exploration_2018,
	title = {Exploration by {Random} {Network} {Distillation}},
	url = {http://arxiv.org/abs/1810.12894},
	doi = {10.48550/arXiv.1810.12894},
	urldate = {2026-02-24},
	publisher = {arXiv},
	author = {Burda, Yuri and Edwards, Harrison and Storkey, Amos and Klimov, Oleg},
	month = oct,
	year = {2018},
	note = {arXiv:1810.12894 [cs]},
}

@inproceedings{pathak_curiosity-driven_2017,
	title = {Curiosity-driven {Exploration} by {Self}-supervised {Prediction}},
	issn = {2640-3498},
	url = {https://proceedings.mlr.press/v70/pathak17a.html},
	language = {en},
	urldate = {2026-02-16},
	booktitle = {Proceedings of the 34th {International} {Conference} on {Machine} {Learning}},
	publisher = {PMLR},
	author = {Pathak, Deepak and Agrawal, Pulkit and Efros, Alexei A. and Darrell, Trevor},
	month = jul,
	year = {2017},
	pages = {2778--2787},
}

@misc{frazier_tutorial_2018,
	title = {A {Tutorial} on {Bayesian} {Optimization}},
	url = {http://arxiv.org/abs/1807.02811},
	doi = {10.48550/arXiv.1807.02811},
	urldate = {2026-02-16},
	publisher = {arXiv},
	author = {Frazier, Peter I.},
	year = {2018},
	note = {arXiv:1807.02811 [stat]},
}

@misc{redmon_yolov3_2018,
	title = {{YOLOv3}: {An} {Incremental} {Improvement}},
	shorttitle = {{YOLOv3}},
	url = {http://arxiv.org/abs/1804.02767},
	doi = {10.48550/arXiv.1804.02767},
	urldate = {2026-02-12},
	publisher = {arXiv},
	author = {Redmon, Joseph and Farhadi, Ali},
	year = {2018},
	note = {arXiv:1804.02767 [cs]},
}

@InProceedings{khodabandeh_robust_2019,
	author = {Khodabandeh, Mehran and Vahdat, Arash and Ranjbar, Mani and Macready, William G.},
	title = {A Robust Learning Approach to Domain Adaptive Object Detection},
	booktitle = {ICCV},
	year = {2019}
}

@article{zhao_object_2019,
	title = {Object {Detection} {With} {Deep} {Learning}: {A} {Review}},
	volume = {30},
	issn = {2162-2388},
	shorttitle = {Object {Detection} {With} {Deep} {Learning}},
	url = {https://ieeexplore.ieee.org/abstract/document/8627998},
	doi = {10.1109/TNNLS.2018.2876865},
	number = {11},
	urldate = {2026-02-16},
	journal = {IEEE Transactions on Neural Networks and Learning Systems},
	author = {Zhao, Zhong-Qiu and Zheng, Peng and Xu, Shou-Tao and Wu, Xindong},
	month = nov,
	year = {2019},
	pages = {3212--3232},
}

@inproceedings{yoo_learning_2019,
	author = {Yoo, Donggeun and Kweon, In So},
	title = {Learning Loss for Active Learning},
	booktitle = {CVPR},
	year = {2019}
}

@inproceedings{carion_end--end_2020,
	title = {End-to-{End} {Object} {Detection} with {Transformers}},
	isbn = {978-3-030-58452-8},
	doi = {10.1007/978-3-030-58452-8_13},
	language = {en},
	booktitle = {ECCV},
	publisher = {Springer International Publishing},
	author = {Carion, Nicolas and Massa, Francisco and Synnaeve, Gabriel and Usunier, Nicolas and Kirillov, Alexander and Zagoruyko, Sergey},
	year = {2020},
}

@article{chaplot_semantic_2020,
	series = {Lecture {Notes} in {Computer} {Science} (including subseries {Lecture} {Notes} in {Artificial} {Intelligence} and {Lecture} {Notes} in {Bioinformatics})},
	title = {Semantic {Curiosity} for {Active} {Visual} {Learning}: 16th {European} {Conference} on {Computer} {Vision}, {ECCV} 2020},
	issn = {9783030585389},
	shorttitle = {Semantic {Curiosity} for {Active} {Visual} {Learning}},
	url = {https://www.scopus.com/pages/publications/85097419988},
	doi = {10.1007/978-3-030-58539-6_19},
	urldate = {2026-02-16},
	journal = {Computer Vision  - ECCV 2020 - 16th European Conference, 2020, Proceedings},
	publisher = {Springer},
	author = {Chaplot, Devendra Singh and Jiang, Helen and Gupta, Saurabh and Gupta, Abhinav},
	editor = {Vedaldi, Andrea and Bischof, Horst and Brox, Thomas and Frahm, Jan-Michael},
	year = {2020},
	pages = {309--326},
}

@article{agarwal2021deep,
  title={Deep reinforcement learning at the edge of the statistical precipice},
  author={Agarwal, Rishabh and Schwarzer, Max and Castro, Pablo Samuel and Courville, Aaron C and Bellemare, Marc},
  journal={Advances in Neural Information Processing Systems},
  volume={34},
  year={2021}
}

@inproceedings{santejudean_path-aware_2021,
	title = {Path-aware optimistic optimization for a mobile robot},
	issn = {2576-2370},
	url = {https://ieeexplore.ieee.org/abstract/document/9683546},
	doi = {10.1109/CDC45484.2021.9683546},
	urldate = {2026-02-16},
	booktitle = {2021 60th {IEEE} {Conference} on {Decision} and {Control} ({CDC})},
	author = {Sântejudean, Tudor and Buşoniu, Lucian},
	year = {2021},
	note = {ISSN: 2576-2370},
}

@inproceedings{chaplot_seal_2021,
	title = {{SEAL}: {Self}-supervised {Embodied} {Active} {Learning} using {Exploration} and {3D} {Consistency}},
	url = {https://proceedings.neurips.cc/paper_files/paper/2021/file/6d0c932802f6953f70eb20931645fa40-Paper.pdf},
	booktitle = {Advances in {Neural} {Information} {Processing} {Systems}},
	publisher = {Curran Associates, Inc.},
	author = {Chaplot, Devendra Singh and Dalal, Murtaza and Gupta, Saurabh and Malik, Jitendra and Salakhutdinov, Russ R},
	year = {2021},
}

@inproceedings{zhou_detecting_2022,
	title = {Detecting {Twenty}-{Thousand} {Classes} {Using} {Image}-{Level} {Supervision}},
	isbn = {978-3-031-20077-9},
	booktitle = {ECCV},
	publisher = {Springer Nature Switzerland},
	author = {Zhou, Xingyi and Girdhar, Rohit and Joulin, Armand and Krähenbühl, Philipp and Misra, Ishan},
	year = {2022},
}

@inproceedings{zheng_towards_2022,
	title = {Towards {Optimal} {Correlational} {Object} {Search}},
	url = {https://ieeexplore.ieee.org/abstract/document/9812252},
	doi = {10.1109/ICRA46639.2022.9812252},
	urldate = {2026-02-16},
	booktitle = {ICRA},
	author = {Zheng, Kaiyu and Chitnis, Rohan and Sung, Yoonchang and Konidaris, George and Tellex, Stefanie},
	year = {2022},
}

@InProceedings{Chen_2018_CVPR,
	author = {Chen, Yuhua and Li, Wen and Sakaridis, Christos and Dai, Dengxin and Van Gool, Luc},
	title = {Domain Adaptive Faster R-CNN for Object Detection in the Wild},
	booktitle = {CVPR},
	year = {2018}
}

@inproceedings{zhai_scaling_2022,
    author    = {Zhai, Xiaohua and Kolesnikov, Alexander and Houlsby, Neil and Beyer, Lucas},
    title     = {Scaling Vision Transformers},
    booktitle = {CVPR},
    year      = {2022},
}

@inproceedings{yu_consistency-based_2022,
    author    = {Yu, Weiping and Zhu, Sijie and Yang, Taojiannan and Chen, Chen},
    title     = {Consistency-Based Active Learning for Object Detection},
    booktitle = {CVPR Workshops},
    year      = {2022},
}

@misc{kolve_ai2-thor_2022,
	title = {{AI2}-{THOR}: {An} {Interactive} {3D} {Environment} for {Visual} {AI}},
	shorttitle = {{AI2}-{THOR}},
	url = {http://arxiv.org/abs/1712.05474},
	doi = {10.48550/arXiv.1712.05474},
	language = {en},
	urldate = {2026-02-12},
	publisher = {arXiv},
	author = {Kolve, Eric and Mottaghi, Roozbeh and Han, Winson and VanderBilt, Eli and Weihs, Luca and Herrasti, Alvaro and Deitke, Matt and Ehsani, Kiana and Gordon, Daniel and Zhu, Yuke and Kembhavi, Aniruddha and Gupta, Abhinav and Farhadi, Ali},
	year = {2022},
	note = {arXiv:1712.05474 [cs]},
}

@misc{deitke_procthor_2022,
	title = {{ProcTHOR}: {Large}-{Scale} {Embodied} {AI} {Using} {Procedural} {Generation}},
	shorttitle = {{ProcTHOR}},
	url = {http://arxiv.org/abs/2206.06994},
	doi = {10.48550/arXiv.2206.06994},
	language = {en},
	urldate = {2026-02-12},
	publisher = {arXiv},
	author = {Deitke, Matt and VanderBilt, Eli and Herrasti, Alvaro and Weihs, Luca and Salvador, Jordi and Ehsani, Kiana and Han, Winson and Kolve, Eric and Farhadi, Ali and Kembhavi, Aniruddha and Mottaghi, Roozbeh},
	year = {2022},
	note = {arXiv:2206.06994 [cs]},
}

@inproceedings{gordon_iqa_2018,
	author = {Gordon, Daniel and Kembhavi, Aniruddha and Rastegari, Mohammad and Redmon, Joseph and Fox, Dieter and Farhadi, Ali},
	title = {IQA: Visual Question Answering in Interactive Environments},
	booktitle = {CVPR},
	year = {2018}
}

@misc{jocher_ultralyticsyolov5_2022,
	title = {ultralytics/yolov5: v7.0 - {YOLOv5} {SOTA} {Realtime} {Instance} {Segmentation}},
	shorttitle = {ultralytics/yolov5},
	url = {https://zenodo.org/records/7347926},
	doi = {10.5281/zenodo.7347926},
	urldate = {2026-02-12},
	publisher = {Zenodo},
	author = {Jocher, Glenn and Chaurasia, Ayush and Stoken, Alex and Borovec, Jirka and NanoCode012 and Kwon, Yonghye and Michael, Kalen and TaoXie and Fang, Jiacong and imyhxy and Lorna and Yifu, Zeng and Wong, Colin and V, Abhiram and Montes, Diego and Wang, Zhiqiang and Fati, Cristi and Nadar, Jebastin and Laughing and UnglvKitDe and Sonck, Victor and tkianai and yxNONG and Skalski, Piotr and Hogan, Adam and Nair, Dhruv and Strobel, Max and Jain, Mrinal},
	year = {2022},
}

@article{minderer2023scaling,
	title = {Scaling {Open}-{Vocabulary} {Object} {Detection}},
	volume = {36},
	url = {https://proceedings.neurips.cc/paper_files/paper/2023/hash/e6d58fc68c0f3c36ae6e0e64478a69c0-Abstract-Conference.html},
	language = {en},
	urldate = {2026-02-23},
	journal = {Advances in Neural Information Processing Systems},
	author = {Minderer, Matthias and Gritsenko, Alexey and Houlsby, Neil},
	month = dec,
	year = {2023},
	pages = {72983--73007},
}

@inproceedings{marza_autonerf_2023,
	title = {{AutoNeRF}: {Training} {Implicit} {Scene} {Representations} with {Autonomous} {Agents}},
	issn = {2153-0866},
	shorttitle = {{AutoNeRF}},
	url = {https://ieeexplore.ieee.org/abstract/document/10802101},
	doi = {10.1109/IROS58592.2024.10802101},
	urldate = {2026-02-16},
	booktitle = {IROS},
	author = {Marza, Pierre and Matignon, Laetitia and Simonin, Olivier and Batra, Dhruv and Wolf, Christian and Chaplot, Devendra S.},
	year = {2024},
	note = {ISSN: 2153-0866},
}

@inproceedings{zheng_system_2023,
	title = {A {System} for {Generalized} {3D} {Multi}-{Object} {Search}},
	url = {https://ieeexplore.ieee.org/abstract/document/10161387},
	doi = {10.1109/ICRA48891.2023.10161387},
	urldate = {2026-02-16},
	booktitle = {ICRA},
	author = {Zheng, Kaiyu and Paul, Anirudha and Tellex, Stefanie},
	year = {2023},
}

@article{ayers_query-based_2023,
	title = {Query-{Based} {Hard}-{Image} {Retrieval} for {Object} {Detection} at {Test} {Time}},
	copyright = {Copyright (c) 2023 Association for the Advancement of Artificial Intelligence},
	issn = {2374-3468},
	url = {https://ojs.aaai.org/index.php/AAAI/article/view/26717},
	doi = {10.1609/aaai.v37i12.26717},
	language = {en},
	number = {12},
	urldate = {2026-02-19},
	journal = {AAAI},
	author = {Ayers, Edward and Sadeghi, Jonathan and Redford, John and Mueller, Romain and Dokania, Puneet K.},
	year = {2023},
}

@misc{pmlr-v205-li23a,
      title={BEHAVIOR-1K: A Human-Centered, Embodied AI Benchmark with 1,000 Everyday Activities and Realistic Simulation}, 
      author={Chengshu Li and Ruohan Zhang and Josiah Wong and Cem Gokmen and Sanjana Srivastava and Roberto Martín-Martín and Chen Wang and Gabrael Levine and Wensi Ai and Benjamin Martinez and Hang Yin and Michael Lingelbach and Minjune Hwang and Ayano Hiranaka and Sujay Garlanka and Arman Aydin and Sharon Lee and Jiankai Sun and Mona Anvari and Manasi Sharma and Dhruva Bansal and Samuel Hunter and Kyu-Young Kim and Alan Lou and Caleb R Matthews and Ivan Villa-Renteria and Jerry Huayang Tang and Claire Tang and Fei Xia and Yunzhu Li and Silvio Savarese and Hyowon Gweon and C. Karen Liu and Jiajun Wu and Li Fei-Fei},
      year={2024},
      eprint={2403.09227},
      archivePrefix={arXiv},
      primaryClass={cs.RO},
      url={https://arxiv.org/abs/2403.09227}, 
}

@misc{zheng_generalized_2023,
	title = {Generalized {Object} {Search}},
	url = {http://arxiv.org/abs/2301.10121},
	doi = {10.48550/arXiv.2301.10121},
	urldate = {2026-02-12},
	publisher = {arXiv},
	author = {Zheng, Kaiyu},
	year = {2023},
	note = {arXiv:2301.10121 [cs]},
}

@inproceedings{yang_plug_2024,
    author    = {Yang, Chenhongyi and Huang, Lichao and Crowley, Elliot J.},
    title     = {Plug and Play Active Learning for Object Detection},
    booktitle = {CVPR},
    year      = {2024},
}

@inproceedings{Majumdar_2024_CVPR,
    author    = {Majumdar, Arjun and Ajay, Anurag and Zhang, Xiaohan and Putta, Pranav and Yenamandra, Sriram and Henaff, Mikael and Silwal, Sneha and Mcvay, Paul and Maksymets, Oleksandr and Arnaud, Sergio and Yadav, Karmesh and Li, Qiyang and Newman, Ben and Sharma, Mohit and Berges, Vincent and Zhang, Shiqi and Agrawal, Pulkit and Bisk, Yonatan and Batra, Dhruv and Kalakrishnan, Mrinal and Meier, Franziska and Paxton, Chris and Sax, Alexander and Rajeswaran, Aravind},
    title     = {OpenEQA: Embodied Question Answering in the Era of Foundation Models},
    booktitle = {CVPR},
    year      = {2024},
}

@inproceedings{liu_grounding_2025,
	title = {Grounding {DINO}: {Marrying} {DINO} with {Grounded} {Pre}-training for {Open}-{Set} {Object} {Detection}},
	isbn = {978-3-031-72970-6},
	shorttitle = {Grounding {DINO}},
	doi = {10.1007/978-3-031-72970-6_3},
	language = {en},
	booktitle = {ECCV},
	author = {Liu, Shilong and Zeng, Zhaoyang and Ren, Tianhe and Li, Feng and Zhang, Hao and Yang, Jie and Jiang, Qing and Li, Chunyuan and Yang, Jianwei and Su, Hang and Zhu, Jun and Zhang, Lei},
	year = {2025},
}

@inproceedings{wu_embodied_2025,
	title = {Embodied {Instruction} {Following} in {Unknown} {Environments}},
	issn = {2153-0866},
	url = {https://ieeexplore.ieee.org/abstract/document/11246127},
	doi = {10.1109/IROS60139.2025.11246127},
	urldate = {2026-02-16},
	booktitle = {IROS},
	author = {Wu, Zhenyu and Wang, Ziwei and Xu, Xiuwei and Yin, Hang and Liang, Yinan and Ma, Angyuan and Lu, Jiwen and Yan, Haibin},
	year = {2025},
	note = {ISSN: 2153-0866},
}

@inproceedings{shi_embodied_2025,
	title = {Embodied {Domain} {Adaptation} for {Object} {Detection}},
	issn = {2153-0866},
	url = {https://ieeexplore.ieee.org/abstract/document/11246195},
	doi = {10.1109/IROS60139.2025.11246195},
	urldate = {2026-02-16},
	booktitle = {IROS},
	author = {Shi, Xiangyu and Qiao, Yanyuan and Liu, Lingqiao and Dayoub, Feras},
	year = {2025},
	note = {ISSN: 2153-0866},
}

@article{munoz_embedded_2025,
	title = {Embedded solution to detect and classify head level objects using stereo vision for visually impaired people with audio feedback},
	copyright = {2025 The Author(s)},
	issn = {2045-2322},
	url = {https://www.nature.com/articles/s41598-025-01529-7},
	doi = {10.1038/s41598-025-01529-7},
	language = {en},
	number = {1},
	urldate = {2026-02-17},
	journal = {Scientific Reports},
	publisher = {Nature Publishing Group},
	author = {Muñoz, Kevin and Chavarria, Mario and Ortiz, Luisa and Suter, Silvan and Schönenberger, Klaus and Bacca-Cortes, Bladimir},
	year = {2025},
}

@article{hong_education_2024,
	title = {Education robot object detection with a brain-inspired approach integrating {Faster} {R}-{CNN}, {YOLOv3}, and semi-supervised learning},
	issn = {1662-5218},
	url = {https://pmc.ncbi.nlm.nih.gov/articles/PMC10794724/},
	doi = {10.3389/fnbot.2023.1338104},
	urldate = {2026-02-17},
	journal = {Frontiers in Neurorobotics},
	author = {Hong, Qing and Dong, Hao and Deng, Wei and Ping, Yihan},
	year = {2024},
}

@InProceedings{pmlr-v270-mirchandani25a,
  title = 	 {So You Think You Can Scale Up Autonomous Robot Data Collection?},
  author =       {Mirchandani, Suvir and Belkhale, Suneel and Hejna, Joey and Choi, Evelyn and Islam, Md Sazzad and Sadigh, Dorsa},
  booktitle = 	 {Conference on Robot Learning (CoRL)},
  year = 	 {2025},
  series = 	 {Proceedings of Machine Learning Research},
  publisher =    {PMLR},
  url = 	 {https://proceedings.mlr.press/v270/mirchandani25a.html},
}

@misc{zhao_detrs_2024,
	title = {{DETRs} {Beat} {YOLOs} on {Real}-time {Object} {Detection}},
	url = {http://arxiv.org/abs/2304.08069},
	doi = {10.48550/arXiv.2304.08069},
	urldate = {2026-02-17},
	publisher = {arXiv},
	author = {Zhao, Yian and Lv, Wenyu and Xu, Shangliang and Wei, Jinman and Wang, Guanzhong and Dang, Qingqing and Liu, Yi and Chen, Jie},
	year = {2024},
	note = {arXiv:2304.08069 [cs]},
}

@article{ucb_paper,
author = {Auer, Peter and Cesa-Bianchi, Nicol\`{o} and Fischer, Paul},
title = {Finite-time Analysis of the Multiarmed Bandit Problem},
year = {2002},
issue_date = {May-June 2002},
publisher = {Kluwer Academic Publishers},
address = {USA},
volume = {47},
number = {2–3},
issn = {0885-6125},
url = {https://doi.org/10.1023/A:1013689704352},
doi = {10.1023/A:1013689704352},
journal = {Mach. Learn.},
month = may,
pages = {235–256},
numpages = {22}
}

@misc{spot,
  author       = {{Boston Dynamics}},
  title        = {{Spot® | Boston Dynamics}},
  year         = {2024},
  url          = {https://www.bostondynamics.com/spot},
  note         = {[Online]}
}
}

\end{document}